\documentclass[runningheads]{llncs}
\usepackage[T1]{fontenc}
\usepackage{graphicx,verbatim}
\usepackage{float}
\usepackage{subfig}
\usepackage{hyperref}       
\usepackage{url}            
\usepackage{booktabs}       
\usepackage{amsfonts}       
\usepackage{nicefrac}       
\usepackage{microtype}      
\usepackage{lipsum}
\usepackage{fancyhdr}       
\graphicspath{{figure/}}     
\usepackage{amsmath}
\usepackage{multirow}
\begin{document}
%
\title{Monocular Depth Guided Occlusion-Aware Disparity Refinement via Semi-supervised Learning in Laparoscopic Images}

\author{Ziteng Liu\inst{1} \and
Dongdong He\inst{1} \and
Chenghong Zhang\inst{1}\and
Wenpeng Gao\inst{1}$^{\star}$\and
Yili Fu\inst{2}$^{\star}$}

\authorrunning{Liu. et al.}
%
\institute{School of Life Science and Technology,  Harbin Institute of Technology, Harbin, China \and
State Key Laboratory of Robotics and System,  Harbin Institute of Technology, Harbin, China }


\maketitle

\begin{abstract}

Occlusion and the scarcity of labeled surgical data are significant challenges in disparity estimation for stereo laparoscopic images. To address these issues, this study proposes a Depth Guided Occlusion-Aware Disparity Refinement Network (DGORNet), which refines disparity maps by leveraging monocular depth information unaffected by occlusion. A Position Embedding (PE) module is introduced to provide explicit spatial context, enhancing the network's ability to localize and refine features.  Furthermore, we introduce an Optical Flow Difference Loss (OFDLoss) for unlabeled data, leveraging temporal continuity across video frames to improve robustness in dynamic surgical scenes.  Experiments on the SCARED dataset demonstrate that DGORNet outperforms state-of-the-art methods in terms of End-Point Error (EPE) and Root Mean Squared Error (RMSE), particularly in occlusion and texture-less regions. Ablation studies confirm the contributions of the Position Embedding and Optical Flow Difference Loss, highlighting their roles in improving spatial and temporal consistency. These results underscore DGORNet's effectiveness in enhancing disparity estimation for laparoscopic surgery, offering a practical solution to challenges in disparity estimation and data limitations.
\end{abstract}

\section{Introduction}


Laparoscopic surgery  remains a clinical gold standard, yet its 2D visualization limits depth perception and anatomical comprehension \cite{Deep2022Cheng,reconstruct2022Yang}. Stereo laparoscopes address this via stereo matching-derived disparity maps for 3D reconstruction \cite{Robust2022Xia}, enhancing intraoperative spatial awareness \cite{stereo2023Wei}. Modern deep disparity estimators leverage end-to-end feature learning \cite{LUO2022Unsupervised}, utilizing cost volume aggregation \cite{psychogyios2022msdesis}. However, occlusion regions caused by the positional disparity between left and right cameras can hinder training and accuracy \cite{DONG2023occlusion}. Therefore, accurately addressing occlusion regions is critical for improving the performance of disparity estimation methods.

To refine disparity maps in occlusion regions, Dong et al. \cite{DONG2023occlusion} classify invalid regions into five types based on the results of the left-right consistency (LRC) check \cite{Zitnick2000LRC} and apply an adaptive disparity reconstruction method tailored to each type. Since occlusions typically occur at the boundaries between different objects, the SDR \cite{Yan2019Segment} refines the disparity map in occlusion regions based on the segmented results. However, these methods often rely on hard thresholds, which can lead to significant estimation errors due to decision inaccuracies \cite{Peng2024OPAL}.
For learning-based methods, incorporating occlusion-aware loss functions is a common strategy to address occlusion issues \cite{Peng2024OPAL,Yuan2024UnSAMFlow}. For instance, Li et al. \cite{Peng2024OPAL} integrate occlusion patterns into the photometric consistency loss \cite{ren2017unsupervised}.
However, these methods fail to resolve the ghosting effect in the warping process \cite{Wang2022OMNET}, which negatively impacts the accuracy of subsequent disparity estimation.

In practice, acquiring ground-truth labels for supervised training is challenging \cite{sun2023scv3}, particularly in laparoscopic surgery, leading to significant data scarcity issues in this domain \cite{Robust2022Xia,Tukra2022Contrastive}. To address this limitation, Psychogyios et al. \cite{psychogyios2022msdesis} propose the MSDESIS framework, which leverages the intrinsic relationship between segmentation and stereo matching to jointly estimate disparity and binary tool segmentation masks. However, the variability of surgical tools across different procedures restricts the generalizability and applicability of this approach. 

Since occlusion is an inherent challenge in disparity estimation tasks, we propose a \textbf{D}epth \textbf{G}uided \textbf{O}cclusion-Aware Disparity \textbf{R}efinement Network (DGORNet) to refine occluded regions in disparity maps using depth information, as monocular depth estimation is not affected by occlusion. Given that occlusion arises from the relative positions of the left and right cameras relative to the target object, we introduce a Position Embedding module to provide additional spatial information, enabling the network to better understand spatial relationships.
To further enhance the utilization of stereo information from image pairs and exploit temporal continuity between video frames, we design an innovative semi-supervised training loss function based on optical flow differential constraints. 
Finally, we evaluate our network against four state-of-the-art disparity estimation methods. The experimental results demonstrate the effectiveness of our network in refining disparity maps.

\section{Method}

In this section, we first introduce the structure of our network. Then the design of the occlusion module is illustrated. Next, we describe the Optical Flow Difference Loss.

\subsection{Network Structure}

    \begin{figure}[!htbp]
		\centering
        \includegraphics[width=\linewidth]{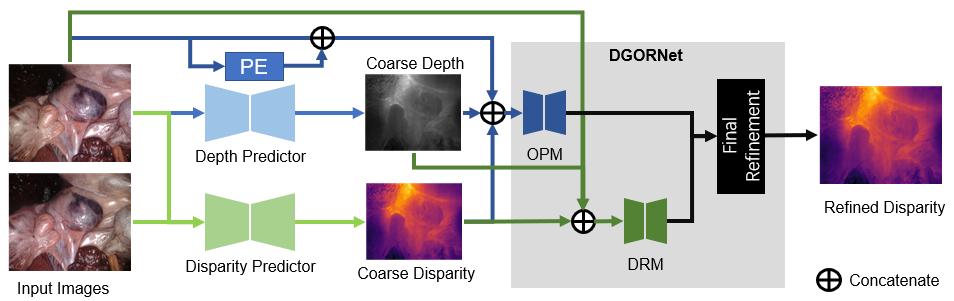}
		\caption{The overview of our network. OPM is the Occlusion Mask Prediction Module. DRM is the Disparity Refinement Module. PE is the Position Embedding. }
		\label{fig:overview}
\end{figure}

The overall architecture of our network is illustrated in Fig. \ref{fig:overview}. The network begins with a depth predictor and a disparity predictor, which generate coarse depth and coarse disparity maps, respectively. The proposed network refines the occlusion regions in the coarse disparity map using guidance from the coarse depth. It comprises two key components: the Occlusion Mask Prediction Module (OPM) and the Disparity Refinement Module (DRM). The OPM is constructed using a sequence of ConvBnLRelu layers \cite{psychogyios2022msdesis}, which progressively refine input features through a series of convolutional operations (64 → 128 → 128 → 64 → 32 → 16 channels). A final convolutional layer outputs the occlusion mask. The OPM takes as input the left image, position map, coarse depth, and coarse disparity, and predicts an occlusion mask. 
The DRM is built using a 2D hourglass module \cite{hourglass2018Chang}. It takes as input the left image, occlusion mask, coarse depth, and coarse disparity, and generates a scale map (K) and a shift map (B). These maps are used to compute a refined inverse depth map $\hat{D}_{inv}$ as follows:
\begin{equation}
    \hat{D}_{inv}=KD_{inv}+B
\end{equation}

In the final refinement step, the coarse disparity is refined using the refined inverse depth map and the occlusion mask:

\begin{equation}
    \hat{S}=MS+(1-M)\hat{D}_{inv}
\end{equation}

where $\hat{S}$ is the final refined disparity, and $M \in[0,1]$ is the occlusion mask, with $M\rightarrow1$ indicating occlusion and $M\rightarrow0$ indicating occlusion-free regions.

    DGORNet integrates these modules to refine disparity maps and predict occlusion masks. First, the input features (coarse disparity, left image, inverse depth, and positional maps) are concatenated and passed through the DGOcclusionRefine module to predict the occlusion mask. Next, the input features (coarse disparity, left image, inverse depth, and occlusion mask) are concatenated and passed through the hourglass2D module to refine the inverse depth map. Finally, the refined disparity map is computed as a weighted combination of the coarse disparity and the refined inverse depth, using the occlusion mask as the weighting factor.

 \subsection{Position Embedding}
 
    \begin{figure}[!htbp]
		\centering
        \resizebox{0.8\linewidth}{!}{
        \subfloat[DeepPruner\cite{scared}]{ \includegraphics[width=0.25\linewidth]{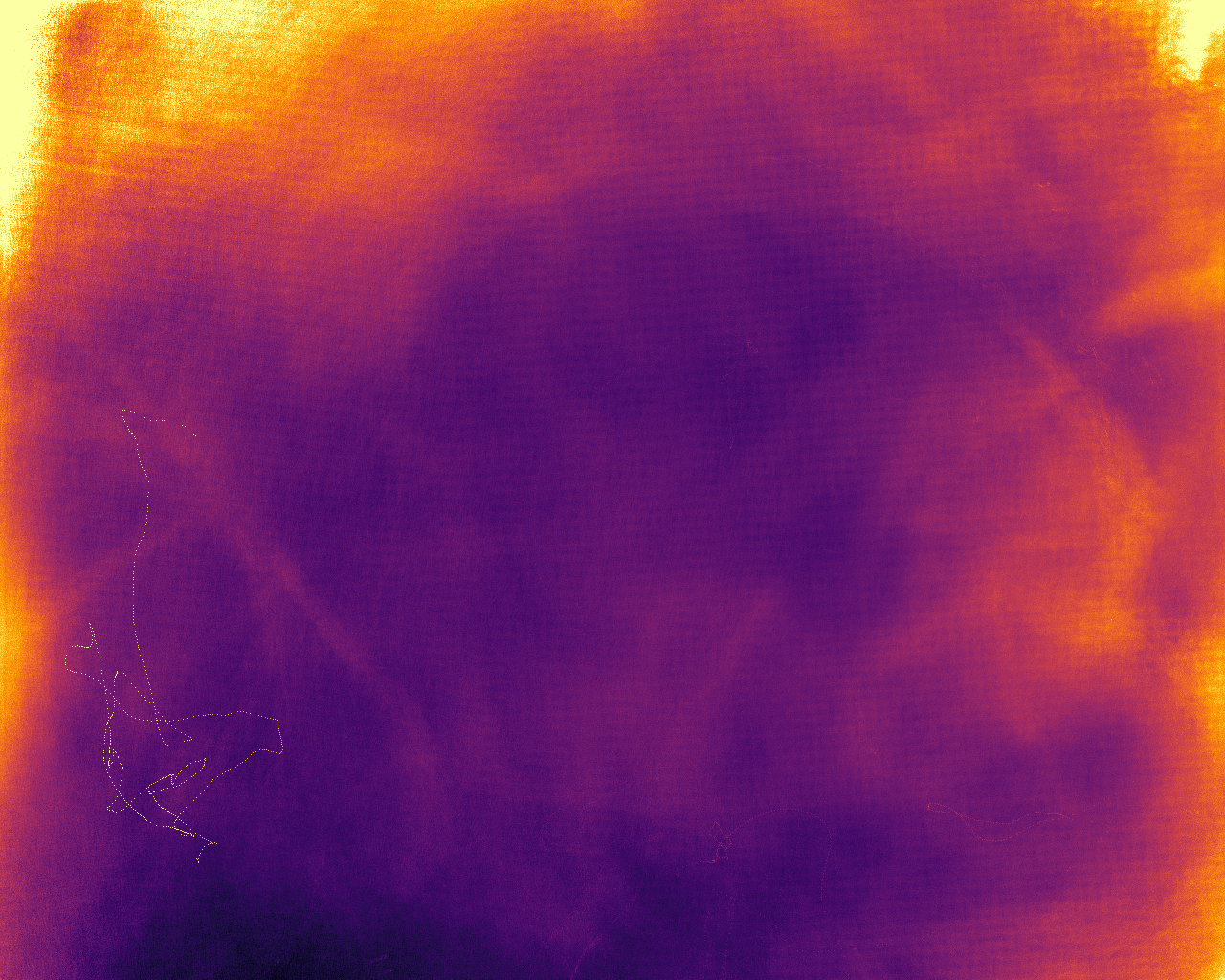}} 
        \subfloat[HSM\cite{yang2019hierarchical}]{\includegraphics[width=0.25\linewidth]{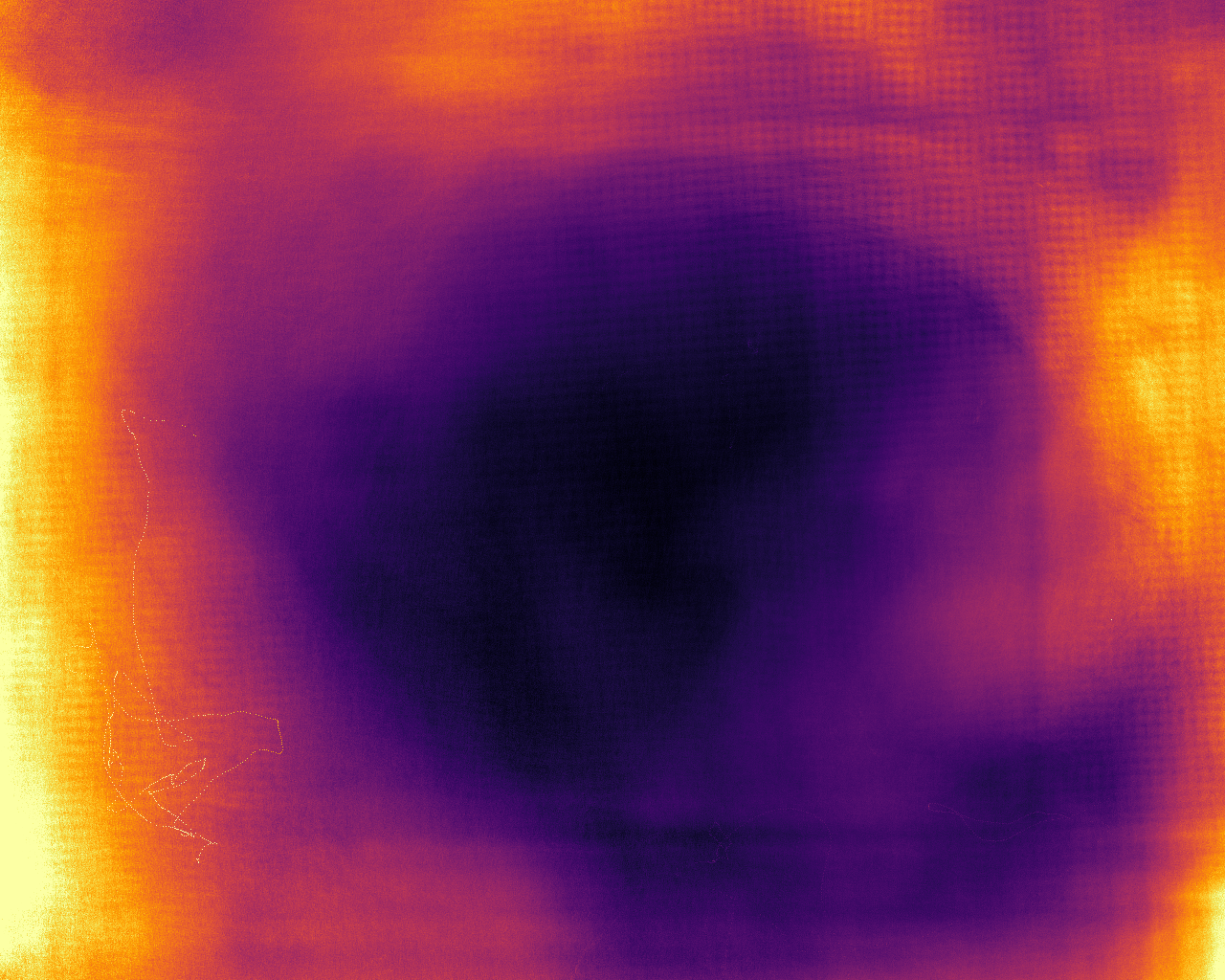}}
        \subfloat[GMStereo\cite{xu2023unimatch}]{ \includegraphics[width=0.25\linewidth]{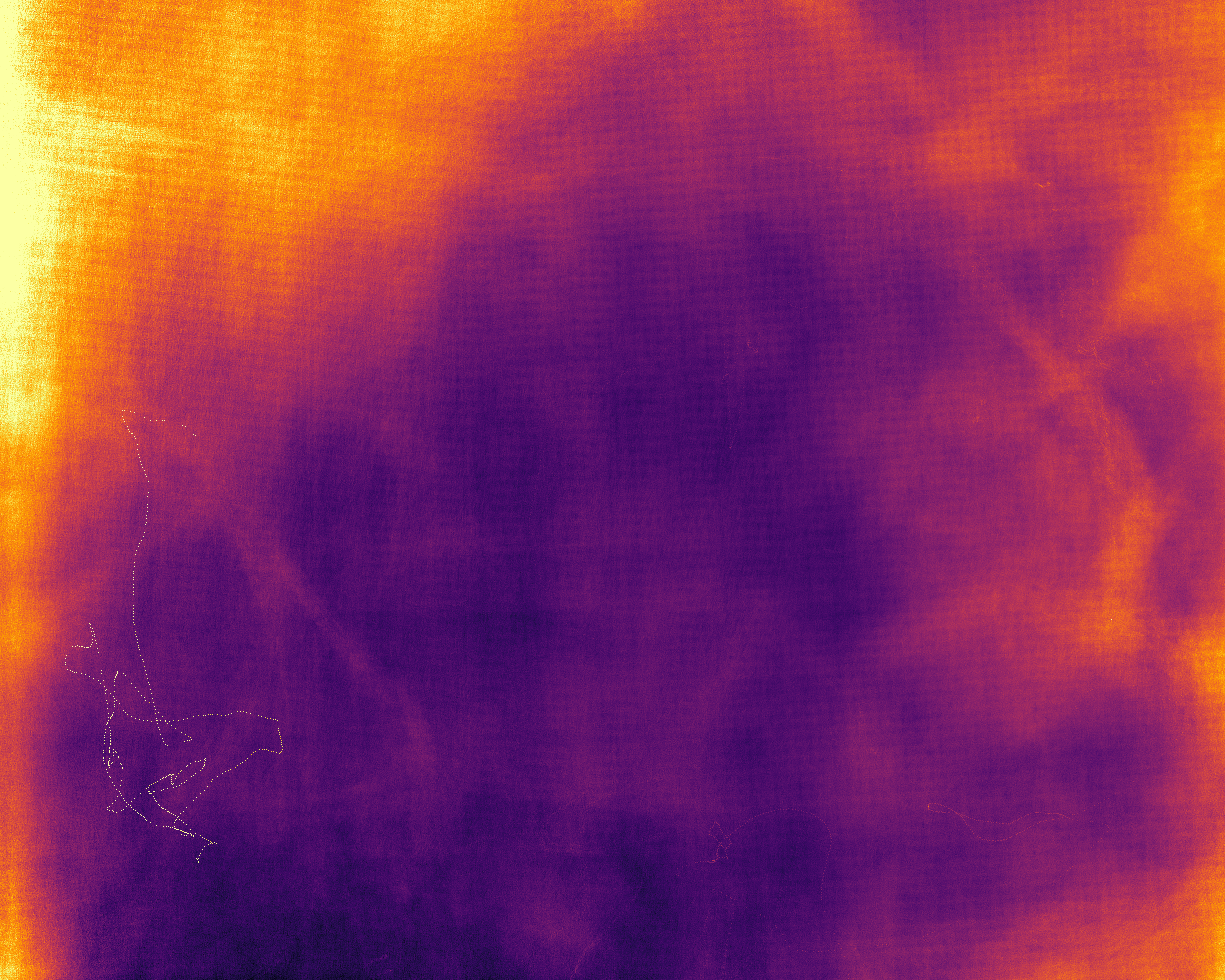}}
        \subfloat[MSDESIS\cite{psychogyios2022msdesis}]{ \includegraphics[width=0.25\linewidth]{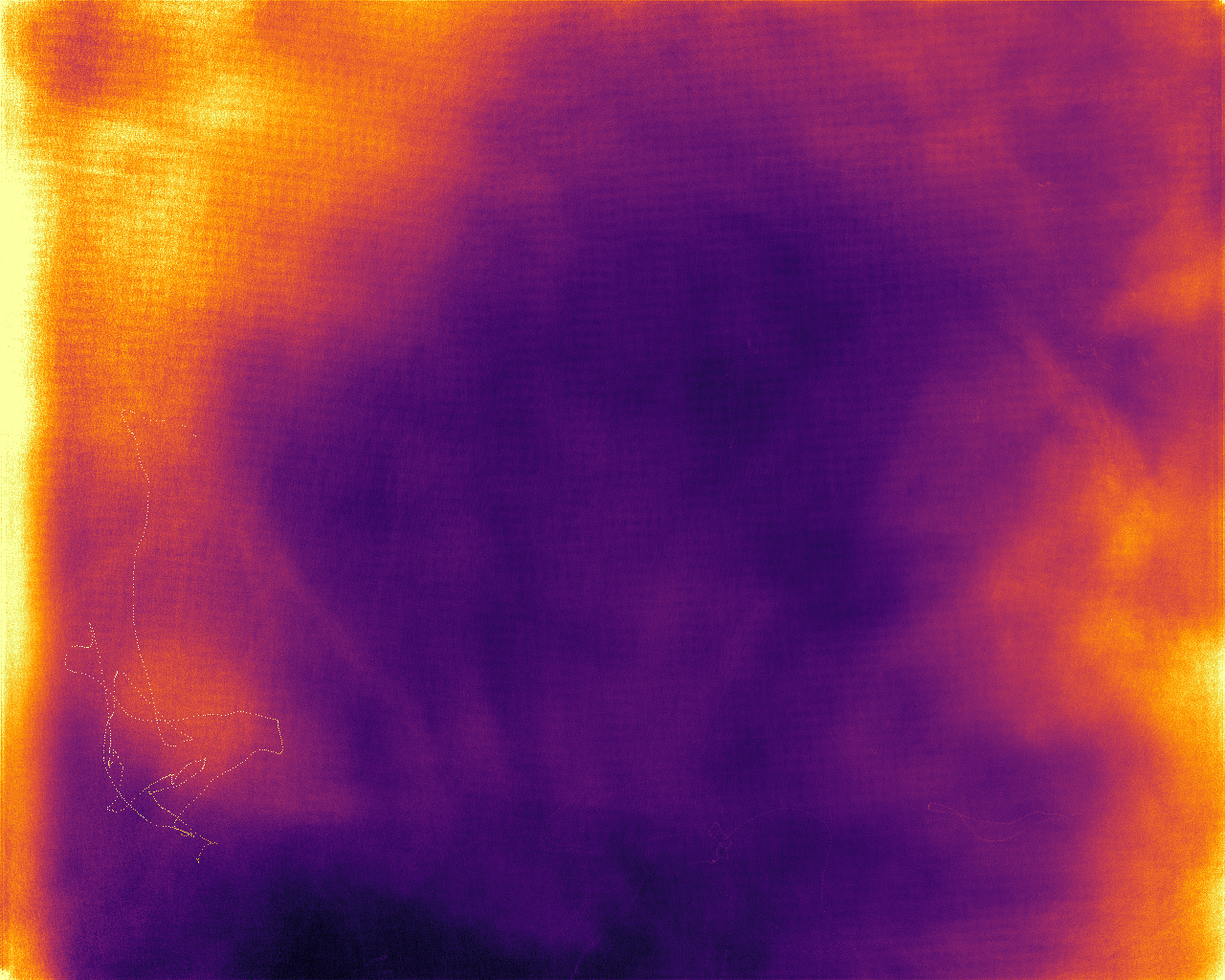}}
        }
		\caption{Error distribution map of disparity prediction result of five state-of-the-art methods on SCARED dataset. Brighter color indicates high error value. 
        }
		\label{fig:err_map}
\end{figure}
    
 As illustrated in Fig. \ref{fig:err_map}, it is evident that the prediction error is correlated with spatial position, a phenomenon also discussed in \cite{DONG2023occlusion}. To address this, a Position Embedding (PE) is incorporated into the OPM by augmenting its input with a position map. This approach generates spatial encodings to provide the network with explicit positional information. Inspired by the position encoding method in Transformers \cite{dosovitskiy2020image}, the PE computes sinusoidal and cosine embeddings for both horizontal and vertical dimensions. These embeddings are concatenated and added to the feature maps, enabling the network to better capture and utilize spatial relationships between pixels. This enhancement is designed to improve the network's ability to localize and refine features, particularly in regions where positional context is critical for accurate predictions.

\subsection{Optical Flow Difference Loss}


In the field of stereo laparoscopic disparity estimation, SCARED dataset is the most widely used dataset providing ground truth disparity. However, due to the constraints of the acquisition environment, the ground truth disparities in SCARED dataset are sparse, and in some video frames, the number of pixels covered by the ground truth is less than $5\%$ of the total. To fully leverage the stereo information from image pairs and the temporal continuity between video frames, this chapter proposes an innovative self-supervised loss for the unlabeled data based on optical flow field differential constraints.

\begin{figure}[!htbp]
		\centering
		\includegraphics[width=0.3\linewidth]{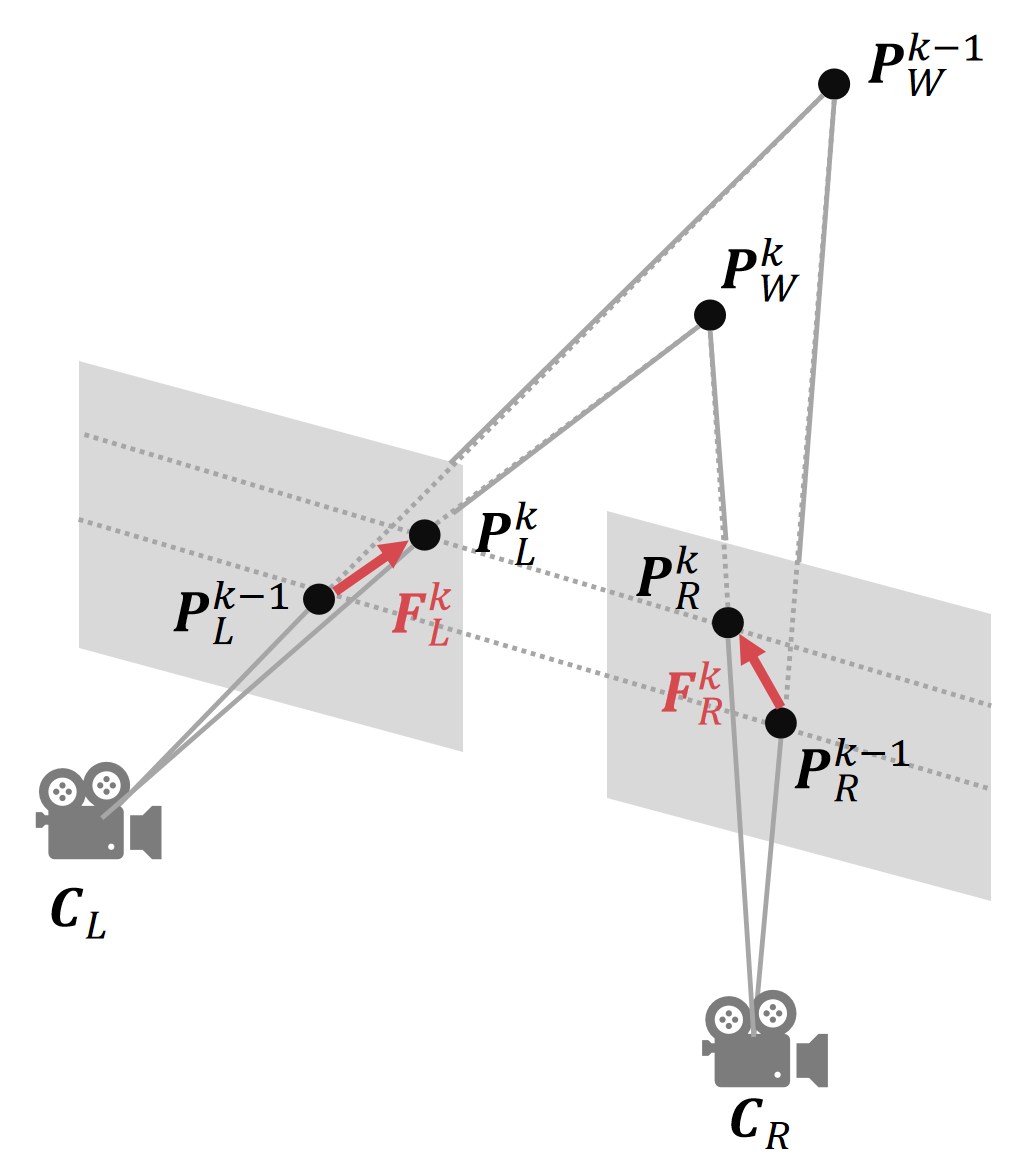}
		\caption{An example of the relationship between a point's movement and disparity.}
		\label{fig:flow_loss}
\end{figure}

As illustrated in Fig. \ref{fig:flow_loss}, $P^{k}_{L}$ and $P^{k-1}_{L}$ represent the positions of point $P_{L}$ in the left image at frames $k$ and $k-1$, respectively. Similarly, $P^{k}_{R}$ and $P^{t-1}_{R}$ denote the positions of point $P_{R}$ in the right image at frames $k$ and $k-1$. The optical flows $\mathbf{F}_{L}^{k}$ and $\mathbf{F}_{R}^{k}$ describe the motion of $P_{L}$ and $P_{R}$ from frame $k-1$ to frame $k$ in the left and right images, respectively. Then, the backward difference between $\mathbf{F}_{L}^{k}$ and $\mathbf{F}_{R}^{k}$ can be expressed as:

\begin{equation}
    \Delta F^k=(P_L^k-P_R^k)-(P_L^{k-1}-P_R^{k-1})
\end{equation}

Under ideal conditions, the y-components of the results on both sides of the equation should consistently be zero. However, in practical applications, deviations in optical flow estimation are unavoidable. To address this, the study utilizes the x-component differences as the loss and the y-component differences as adaptive weights for the loss. Consequently, the Optical Flow Difference Loss (OFDLoss) is defined as:



\begin{equation}
\label{eq:l_flow}
    L_{flow}=w_{flow}\cdot\left ( \Delta_x F^k- \Delta_x^kP \right ) 
\end{equation}

where $\cdot$ represents the Hadamard Product,$\Delta_x$ denotes the x-component difference, and $\Delta_x^kP=\Delta_x P^k-\Delta_x P^{k-1}$. The adaptive loss weight $w_{flow}$ is calculated based on the y-component difference as follows:

\begin{equation}
    w_{flow}=\text{clamp}_{0,1}\left( 1 - \left( \Delta_y F^k \right)^2 \right)
\end{equation}

Here, $\text{clamp}_{0,1}$ ensures that $w_{flow}$ remains within the range [0, 1]. This formulation allows the network to focus more on areas with accurate optical flow estimates while reducing the influence of less reliable predictions.


\section{Experiments}

\subsection{Datasets and Metrics}

The publicly available Stereo Correspondence and Reconstruction of Endoscopic Data (SCARED) dataset \cite{scared} is utilized to evaluate the performance of the proposed model and compare it with other methods. Due to errors in the calibration parameters of datasets 4 and 5 \cite{scared}, datasets 1, 2, 3, 6, and 7 are employed for training, while datasets 8 and 9 are reserved for evaluation. Notably, since the RGB video and the interpolated ground truth are not perfectly synchronized in time \cite{psychogyios2022msdesis,Deep2022Cheng}, a split evaluation dataset is constructed using only the keyframes from datasets 8 and 9. For clarity, the evaluation dataset using the full video is denoted as SCARED-evaluation-full (SEF), and the evaluation dataset using keyframes is denoted as SCARED-evaluation-keyframe (SEK). It should be mentioned that the samples included in SEK are not included in SEF. In summary, the training dataset consists of 17,231 samples, SEF contains 5,915 samples, and SEK contains 10 samples.


For stereo matching, the evaluation metrics include the average end-point error (EPE), defined as the mean absolute error between the reference and predicted disparity values, and the percentage of pixels with a disparity error greater than 3 pixels (Bad3). To evaluate depth prediction performance, we employ the widely used root mean squared error (RMSE).

\subsection{Implementation Details}

The network is implemented using PyTorch and trained on a single NVIDIA A10 GPU. Following established practices \cite{sun2023scv3}, we employ LeReS as the depth predictor. For the disparity predictor, we utilize MSDESIS \cite{psychogyios2022msdesis} and GMStereo \cite{xu2023unimatch} to construct our small and large prototypes, respectively. For calculating optical flow for OFDLoss, this study adopts the widely recognized GMFlow  \cite{xu2023unimatch}. The AdamW optimizer is adopted with a weight decay of \(1 \times 10^{-4}\). The network is trained for 10 epochs with a batch size of 6. A cosine annealing learning rate scheduler is applied, starting with an initial learning rate of \(1 \times 10^{-4}\).

\subsection{Loss Function}
In this study, our network is trained using a semi-supervised loss function defined as follows:

\begin{equation}
\begin{aligned}
        L_{ss}=&L_{s}(\hat{S},S_{gt}) + L_{s}(\hat{D}_{inv},S_{gt}) + 0.5\cdot L_{flow}(\hat{S})
        \\&+ 0.25\cdot L_{DC}(M,M_{LRC})+ 0.25\cdot L_{WBCE}(M,M_{LRC})
\end{aligned}
\end{equation}


where $L_{s}(\hat{S},S_{gt})$ and $L_{s}(\hat{D}_{inv},S_{gt})$ represent the L1 loss between the refined disparity (\(\hat{S}\)) and the ground-truth disparity (\(S_{gt}\)), and between the refined inverse depth (\(\hat{D}_{inv}\)) and the ground-truth disparity, respectively. \( L_{flow}(\hat{S}) \) denotes the proposed optical flow difference loss (Eq. \ref{eq:l_flow}). Additionally, \( L_{DC} \) and \( L_{WBCE} \) correspond to the Dice loss and the weighted binary cross-entropy loss, respectively. The ground-truth occlusion mask (\( M_{LRC} \)) is derived using the Left-Right Consistency (LRC) check \cite{DONG2023occlusion}.

\section{Results and Discussion}

\subsection{Quantitative Comparisons}

 \begin{figure}[!htbp]
		\centering
        \subfloat[Input Image]{ \includegraphics[width=0.21\linewidth]{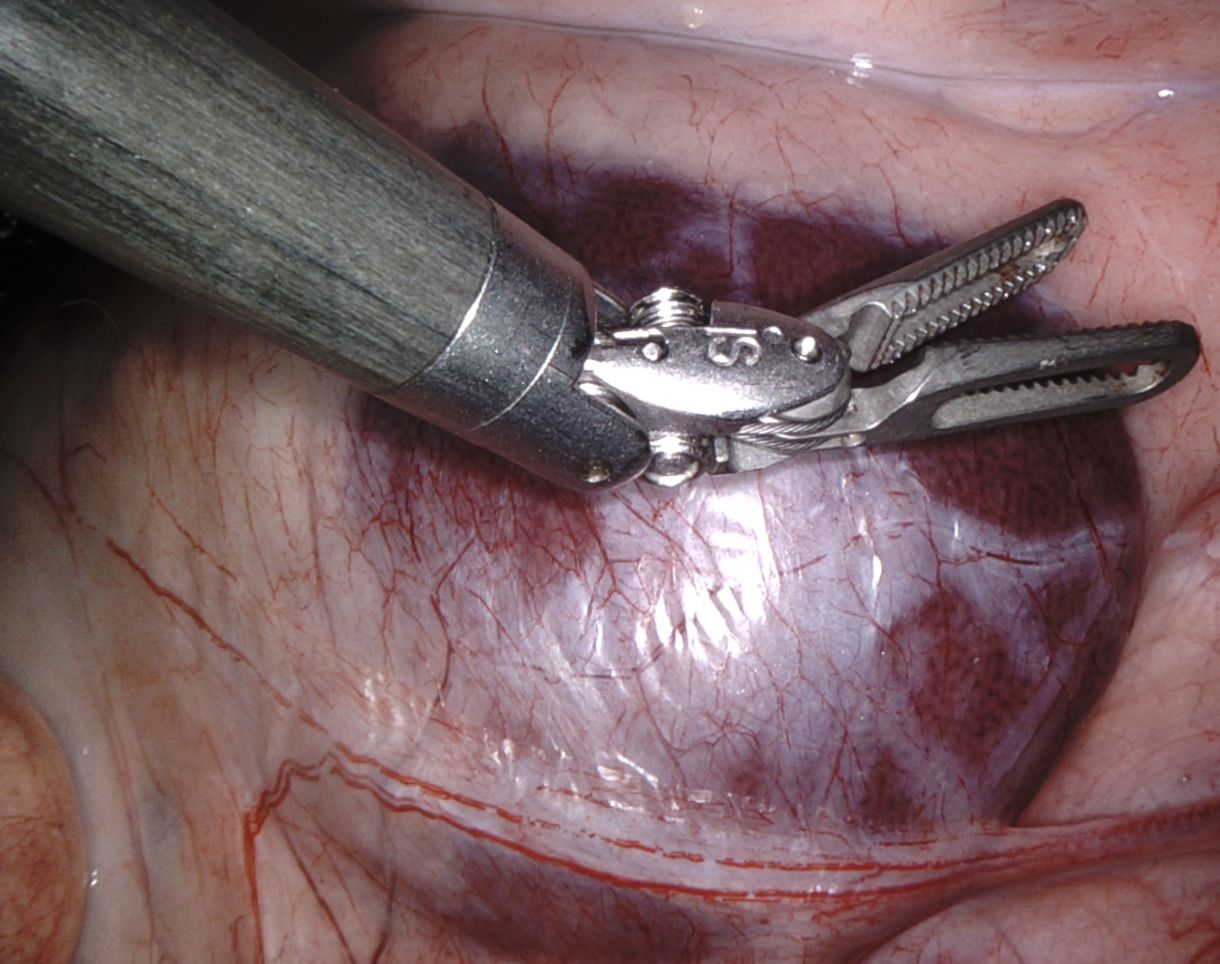}} 
        \subfloat[LeRes]{\includegraphics[width=0.21\linewidth]{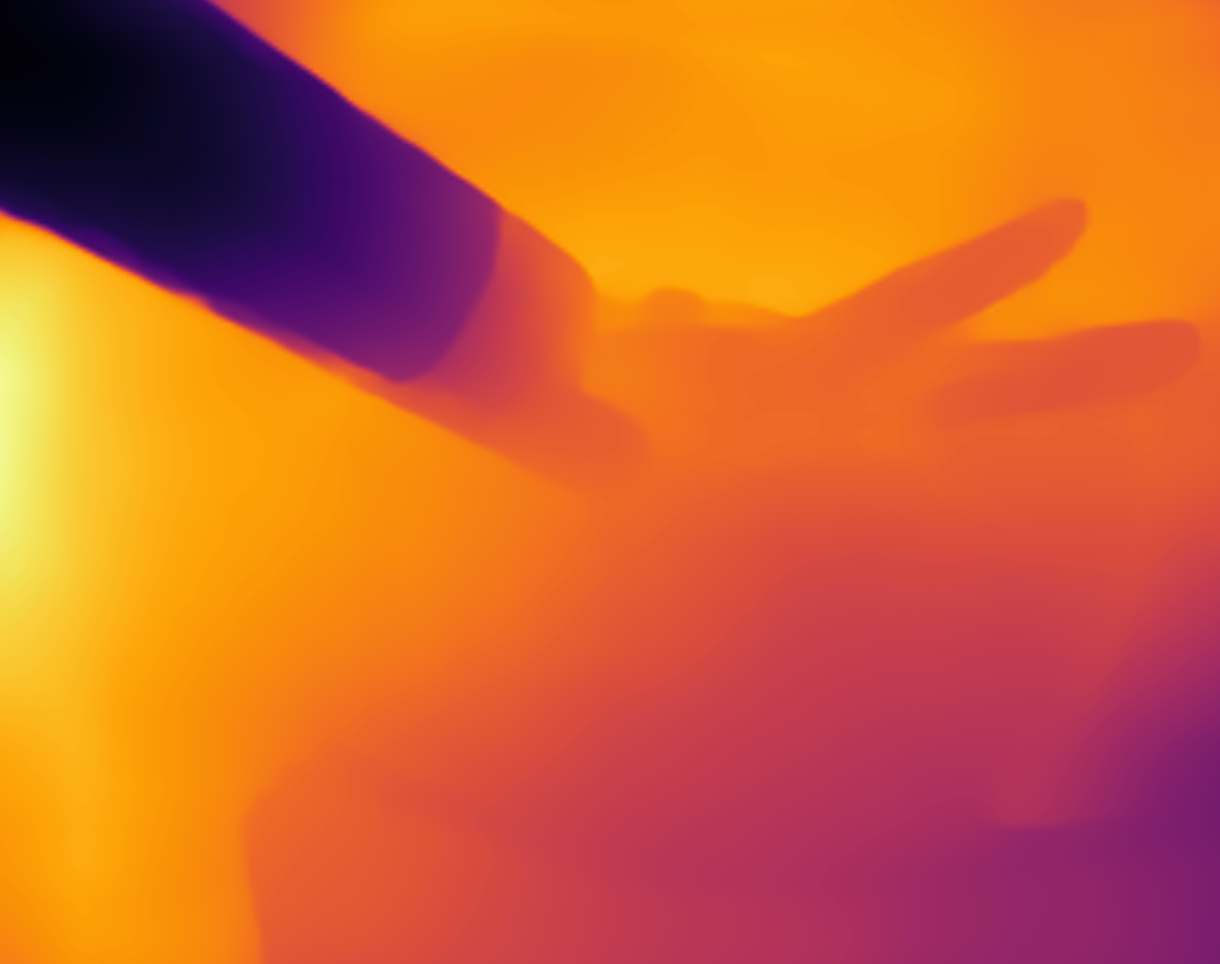}}
        \subfloat[MSDESIS]{ \includegraphics[width=0.21\linewidth]{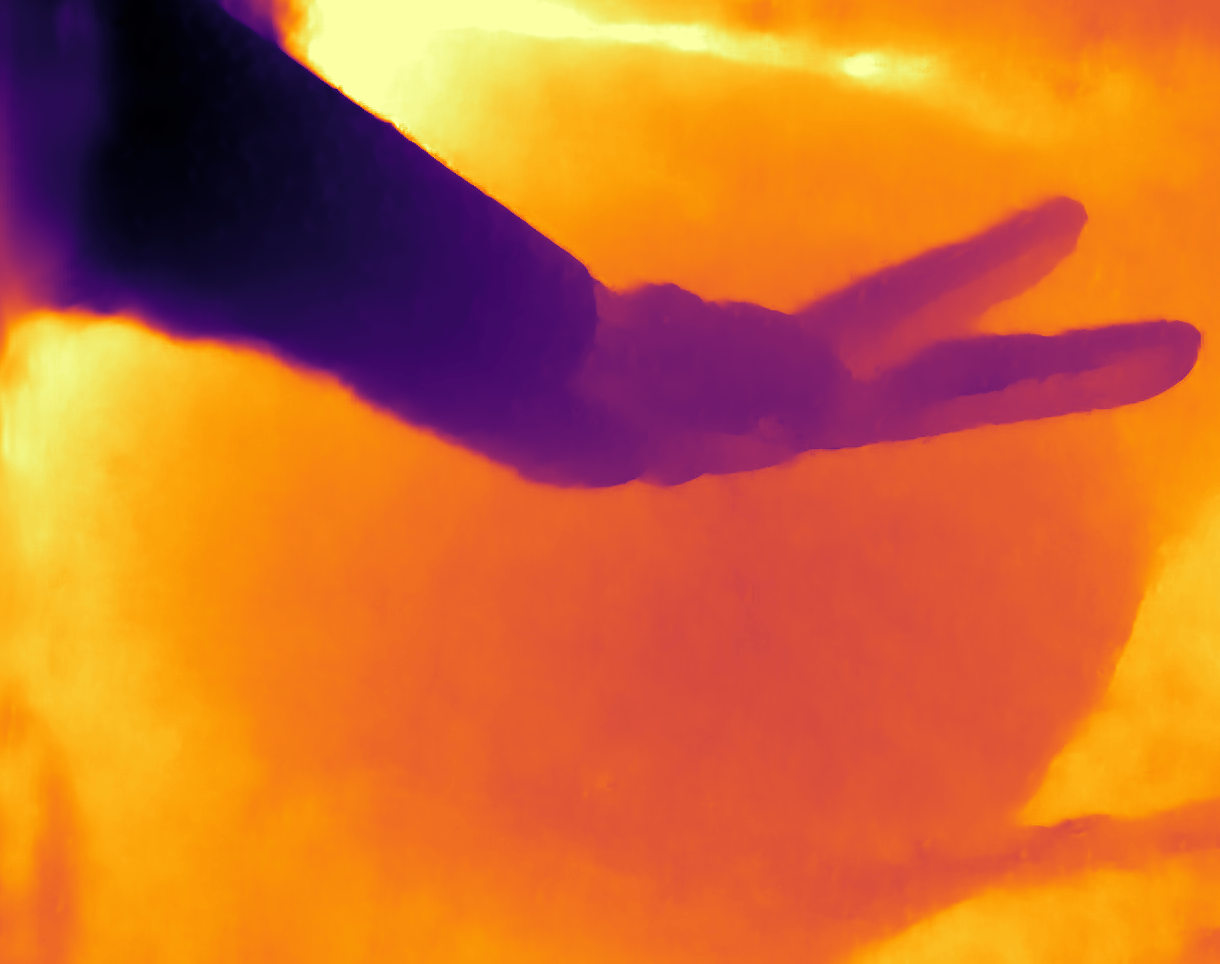}} 
         \subfloat[GMStereo]{ \includegraphics[width=0.21\linewidth]{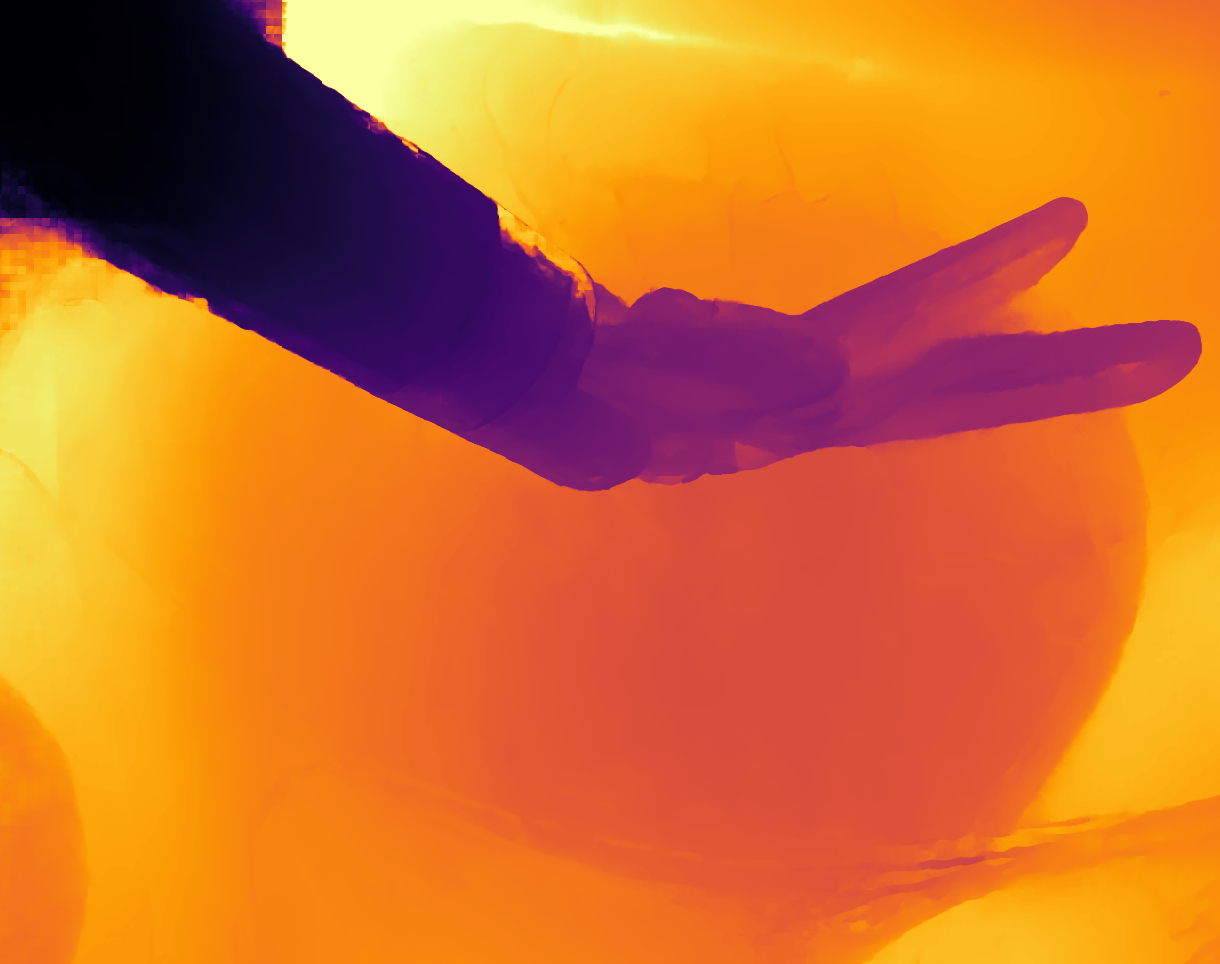}} 
        \quad
        \subfloat[GT]{ \includegraphics[width=0.21\linewidth]{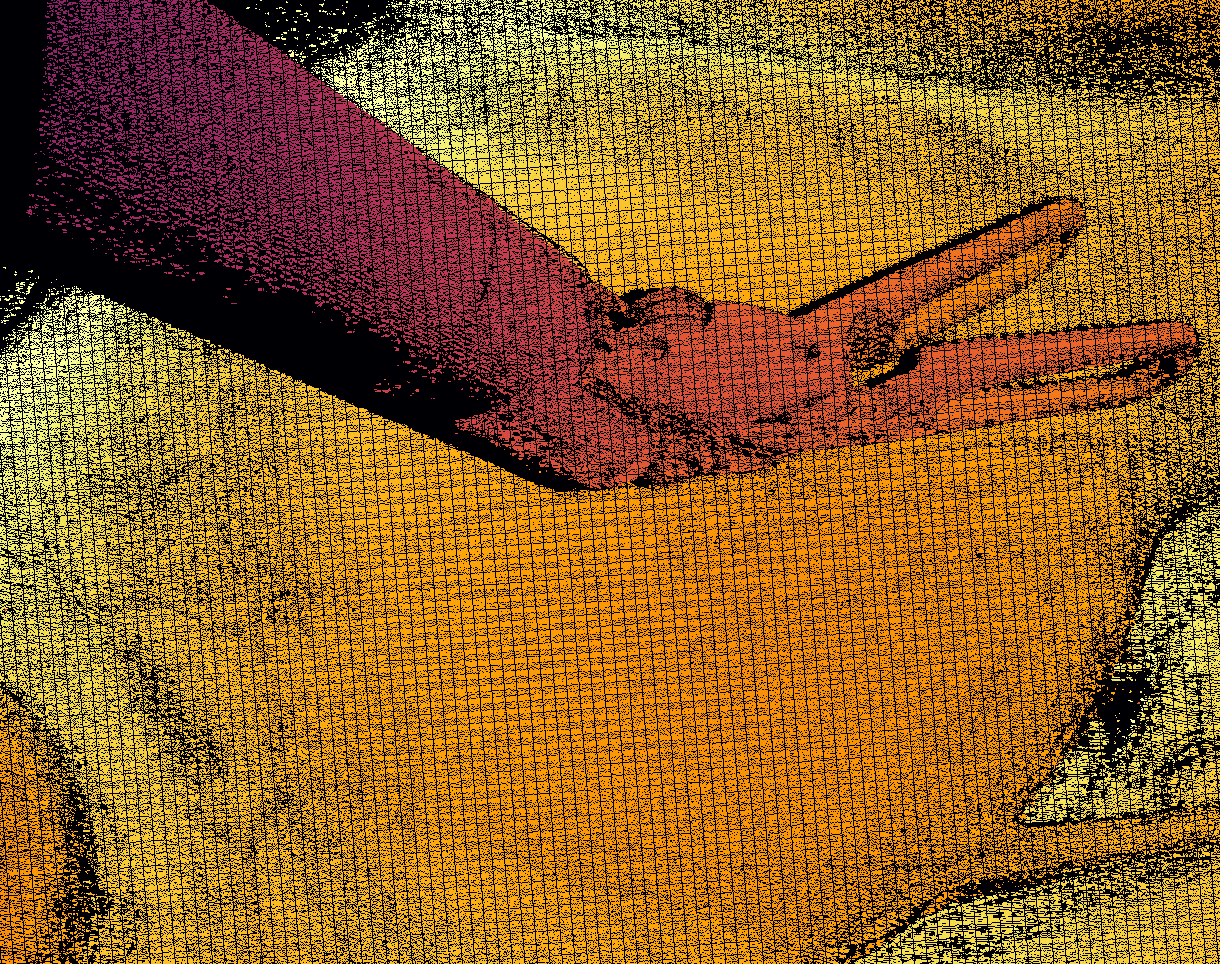}}
        \subfloat[DeepPruner]{ \includegraphics[width=0.21\linewidth]{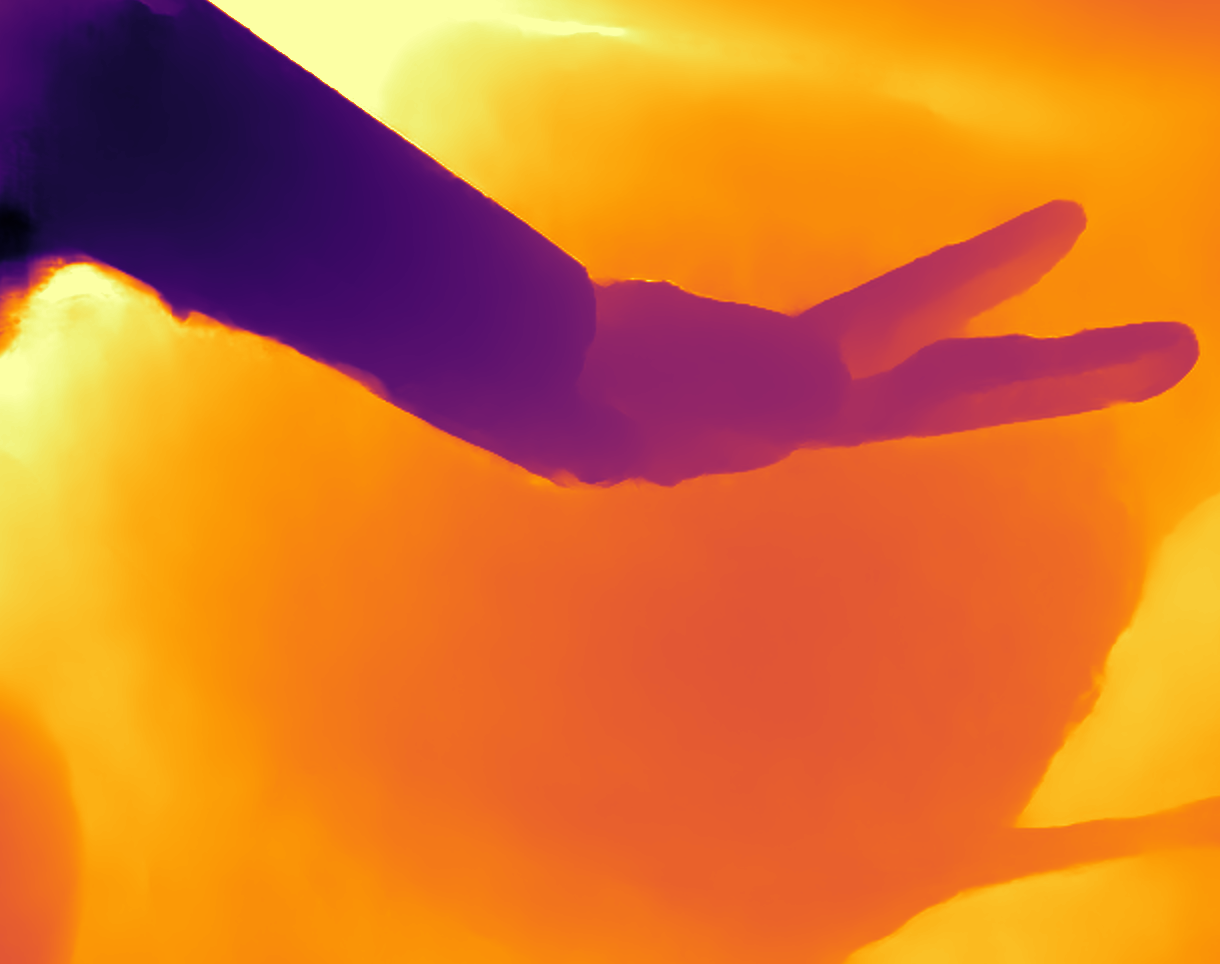}}
        \subfloat[DGOR-LM]{ \includegraphics[width=0.21\linewidth]{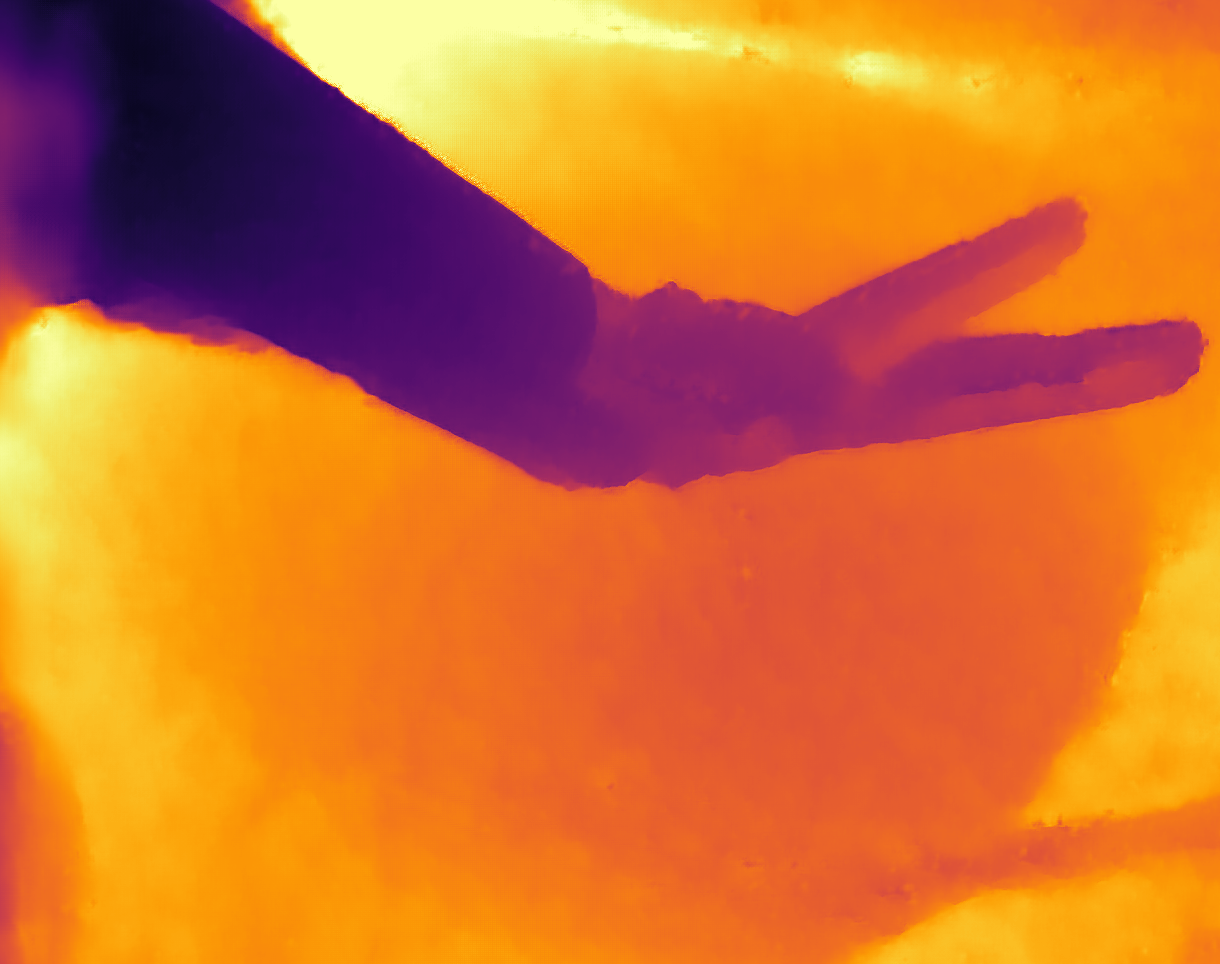}}
        \subfloat[ DGOR-LG]{ \includegraphics[width=0.21\linewidth]{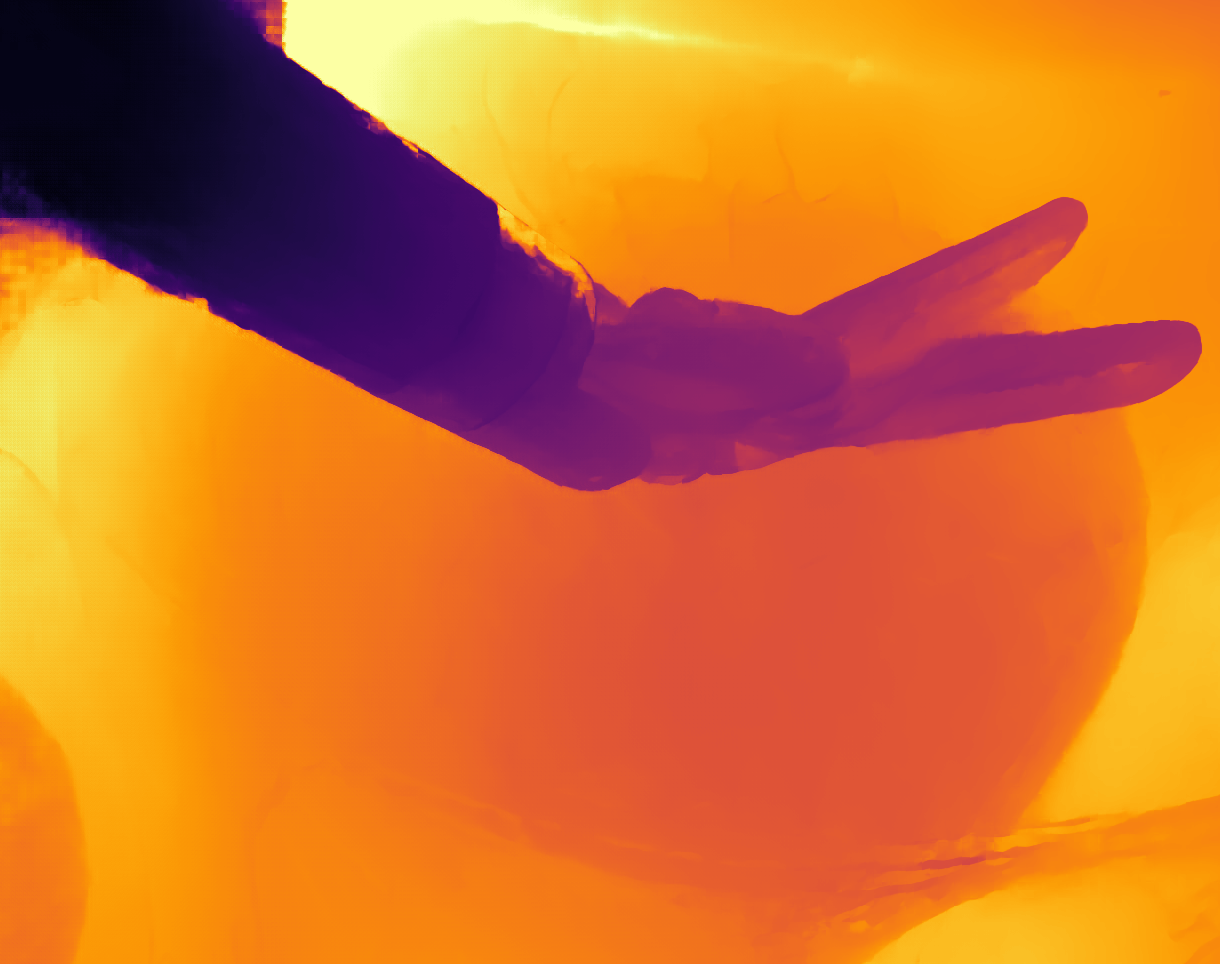}}
		\caption{Prediction result on the SEK. }
		\label{fig:results}
\end{figure}

The evaluation results on the SEF and SEK datasets are presented in Table \ref{tab:results}. Our DGOR-LG model achieves the best EPE results on both datasets, while both the DGOR-LM and DGOR-LG models outperform the baseline disparity predictors (MSDESIS and GMStereo). However, the Bad3 metric of our method is slightly higher than that of the disparity predictors on the SEF dataset. This can be attributed to potential outliers introduced by the depth predictor (LeReS), which may affect the final disparity refinement. 
In terms of depth accuracy, our method outperforms the disparity predictors on the RMSE metric for the SEK dataset, with DGOR-LG achieving the second lowest RMSE. As illustrated in Fig. \ref{fig:results}, the DGOR-LM model generates fewer outliers compared to MSDESIS, particularly in tool edge regions and image borders. 

The refinement modules incur minimal computational overhead (3.83 seconds/1000 predictions), though total runtime depends on base disparity predictors (40.42s for DGOR-LM vs. 207.06s for DGOR-LG). DGOR-LM achieves real-time capability (40.42s) while maintaining competitive accuracy, making it preferable for latency-sensitive applications. This flexibility allows our framework to adapt to different operational constraints without architectural modifications.

These results highlight the effectiveness of our proposed disparity refinement approach, which leverages monocular depth information to refine occlusion areas, thereby improving overall performance in challenging regions.

\begin{table}[]
\centering
\caption{Quantitative results on SEF and SEK. The runtime is measured as the time it takes for a network to make 1000 predictions on 1280 × 1024 inputs. \textbf{BOLD} and \underline{UNDERLINE} indicate the best and the second-best performance, respectively.}
\label{tab:results}
\resizebox{\linewidth}{!}{
\begin{tabular}{llllllllllll}
\hline
Method                                                             &  & \multicolumn{3}{l}{SEF}                                                                                                                                                                &  & \multicolumn{3}{l}{SEK}                                                                                                                                                                &  & \multirow{2}{*}{\begin{tabular}[c]{@{}l@{}}Params\\ {[}millions{]}\end{tabular}} & \multirow{2}{*}{\begin{tabular}[c]{@{}l@{}}Runtime\\ {[}seconds{]}\end{tabular}} \\ \cline{3-5} \cline{7-9}
                                                                   &  & \begin{tabular}[c]{@{}l@{}}EPE\\ {[}pixels{]}\end{tabular} & \begin{tabular}[c]{@{}l@{}}Bad3\\ {[}pixels$\%${]}\end{tabular} & \begin{tabular}[c]{@{}l@{}}RMSE\\ {[}mm{]}\end{tabular} &  & \begin{tabular}[c]{@{}l@{}}EPE\\ {[}pixels{]}\end{tabular} & \begin{tabular}[c]{@{}l@{}}Bad3\\ {[}pixels$\%${]}\end{tabular} & \begin{tabular}[c]{@{}l@{}}RMSE\\ {[}mm{]}\end{tabular} &  &                                                                                  &                                                                                  \\ \hline
LeRes                                                              &  & N/A                                                        & N/A                                                             & 9.33                                                    &  & N/A                                                        & N/A                                                             & 7.31                                                    &  & 52.13                                                                            & 19.58                                                                            \\
HSM                                                                &  & 6.05                                                       & 61.85                                                           & 6.85                                                    &  & 4.92                                                       & 57.00                                                           & 6.14                                                    &  & 3.17                                                                             & 40.71                                                                            \\
DeepPruner                                                         &  & 4.08                                                       & 46.45                                                           & 5.06                                                    &  & \textbf{1.89}                                                       & \textbf{11.67}                                                           & 3.69                                                    &  & 7.49                                                                             & 168.36                                                                           \\
GMStereo                                                           &  & 4.04                                                       & \textbf{43.95}                                                           & 5.03                                                    &  & 1.95                                                       & 13.93                                                           & \textbf{3.33 }                                                   &  & 4.68                                                                             & 183.65                                                                           \\
MSDESIS                                                            &  & 4.45                                                       & 46.04                                                           & 5.40                                                    &  & 2.30                                                       & 13.50                                                           & 3.99                                                    &  & 3.03                                                                             & 17.01                                                                            \\
\begin{tabular}[c]{@{}l@{}}DGOR-LM\\ (LeRes+MSDESIS)\end{tabular}  &  & 4.11                                                       & 47.04                                                           & 5.36                                                    &  & 2.20                                                       & \underline{13.30}                                                           & 3.91                                                    &  & \begin{tabular}[c]{@{}l@{}}0.33\\ (55.49)\end{tabular}                           & \begin{tabular}[c]{@{}l@{}}3.83\\ (40.42)\end{tabular}                           \\
\begin{tabular}[c]{@{}l@{}}DGOR-LG\\ (LeRes+GMStereo)\end{tabular} &  & \textbf{4.02}                                                       & \underline{45.37}                                                           & \textbf{4.93}                                                    &  & \underline{1.94}                                                       & 14.05                                                           & \underline{3.36}                                                    &  & \begin{tabular}[c]{@{}l@{}}0.33\\ (57.15)\end{tabular}                           & \begin{tabular}[c]{@{}l@{}}3.83\\ (207.06)\end{tabular}                          \\ \hline
\end{tabular}
}
\end{table}

\subsection{Ablation Study}
 To validate the effectiveness of the proposed method, ablation studies are conducted to assess the contributions of the Position Embedding and the optical flow difference loss. Both experiments utilize the small prototype (LeReS + MSDESIS) and are evaluated on the SEK dataset.

\begin{itemize}
\item {\textbf{Position Embedding.}  As shown in Table \ref{tab:ab_exp}, removing the Position Embedding module results in a slight degradation in performance across all metrics.  This demonstrates that the Position Embedding module provides valuable spatial context, enhancing the network's ability to localize and refine features. The explicit positional information helps the network better understand the geometric relationships between pixels, leading to improved disparity estimation accuracy. }
\item {\textbf{Optical Flow Difference Loss.} As shown in Table \ref{tab:ab_exp}, removing the Optical Flow Difference Loss leads to a marginal decline in performance.  This suggests that the Optical Flow Difference Loss contributes to refining the temporal consistency of disparity predictions. This loss function capitalizes on motion patterns derived from consecutive video frames, effectively constraining disparity through temporal continuity priors to improve the overall robustness of the disparity refinement process.}

\end{itemize}

\begin{table}[]
\centering
\caption{Ablation study on Position Embedding (PE) and Optical Flow Difference Loss (OFDLoss).}
\label{tab:ab_exp}
\begin{tabular}{lllll}
\hline
Method      &  & EPE   & Bad3  & RMSE  \\ \hline
DGOR-LM     &  & 2.204 & 13.30 & 3.971 \\
w/o PE      &  &   2.243     &   13.85    &   3.984    \\
w/o OFDLoss &  &   2.207   &    13.29   &   3.995    \\ \hline
\end{tabular}

\end{table}


\section{Conclusion}
This study presents DGORNet, a novel framework for refining occluded regions in stereo laparoscopic disparity estimation. The network combines depth guidance from monocular estimation with spatial encoding through position embedding, effectively addressing occlusion challenges inherent in stereo matching. Our semi-supervised optical flow differential loss leverages temporal continuity to enhance consistency without dense annotations.
Experiments on the SCARED dataset demonstrate DGORNet's superiority, achieving state-of-the-art EPE (4.02 pixels) and RMSE (4.93 mm) on SEF. Ablation studies confirm the contributions of PE and OFDLoss.


\bibliographystyle{splncs04}
\bibliography{references}

\begin{thebibliography}{10}
\providecommand{\url}[1]{\texttt{#1}}
\providecommand{\urlprefix}{URL }
\providecommand{\doi}[1]{https://doi.org/#1}

\bibitem{scared}
Allan, M., McLeod, A.J., Wang, C., Rosenthal, J., Hu, Z., Gard, N., Eisert, P., Fu, K.X., Zeffiro, T., Xia, W., Zhu, Z., Luo, H., Jia, F., Zhang, X., Li, X., Sharan, L., Kurmann, T., Schmid, S., Sznitman, R., Psychogyios, D., Azizian, M., Stoyanov, D., Maier{-}Hein, L., Speidel, S.: Stereo correspondence and reconstruction of endoscopic data challenge. CoRR  \textbf{abs/2101.01133} (2021), \url{https://arxiv.org/abs/2101.01133}

\bibitem{hourglass2018Chang}
Chang, J.R., Chen, Y.S.: Pyramid stereo matching network. In: 2018 IEEE/CVF Conference on Computer Vision and Pattern Recognition. pp. 5410--5418 (2018). \doi{10.1109/CVPR.2018.00567}

\bibitem{Deep2022Cheng}
Cheng, X., Zhong, Y., Harandi, M., Drummond, T., Wang, Z., Ge, Z.: Deep laparoscopic stereo matching with transformers. In: Wang, L., Dou, Q., Fletcher, P.T., Speidel, S., Li, S. (eds.) Medical Image Computing and Computer Assisted Intervention -- MICCAI 2022. pp. 464--474. Springer Nature Switzerland, Cham (2022)

\bibitem{DONG2023occlusion}
Dong, L., Han, Y., Hu, M., Luo, H., Wang, Y.: Stereo matching method based on high-precision occlusion-recovering and discontinuity-preserving. Displays  \textbf{80},  102573 (2023). \doi{https://doi.org/10.1016/j.displa.2023.102573}, \url{https://www.sciencedirect.com/science/article/pii/S014193822300207X}

\bibitem{dosovitskiy2020image}
Dosovitskiy, A., Beyer, L., Kolesnikov, A., Weissenborn, D., Zhai, X., Unterthiner, T., Dehghani, M., Minderer, M., Heigold, G., Gelly, S., Uszkoreit, J., Houlsby, N.: An image is worth 16x16 words: Transformers for image recognition at scale. In: International Conference on Learning Representations 2021 (2021), \url{https://openreview.net/forum?id=YicbFdNTTy}

\bibitem{Peng2024OPAL}
Li, P., Zhao, J., Wu, J., Deng, C., Han, Y., Wang, H., Yu, T.: Opal: Occlusion pattern aware loss for unsupervised light field disparity estimation. IEEE Transactions on Pattern Analysis and Machine Intelligence  \textbf{46}(2),  681--694 (2024). \doi{10.1109/TPAMI.2023.3296600}

\bibitem{LUO2022Unsupervised}
Luo, H., Wang, C., Duan, X., Liu, H., Wang, P., Hu, Q., Jia, F.: Unsupervised learning of depth estimation from imperfect rectified stereo laparoscopic images. Computers in Biology and Medicine  \textbf{140},  105109 (2022). \doi{https://doi.org/10.1016/j.compbiomed.2021.105109}, \url{https://www.sciencedirect.com/science/article/pii/S0010482521009033}

\bibitem{psychogyios2022msdesis}
Psychogyios, D., Mazomenos, E., Vasconcelos, F., Stoyanov, D.: Msdesis: Multitask stereo disparity estimation and surgical instrument segmentation. IEEE transactions on medical imaging  \textbf{41}(11),  3218--3230 (2022)

\bibitem{ren2017unsupervised}
Ren, Z., Yan, J., Ni, B., Liu, B., Yang, X., Zha, H.: Unsupervised deep learning for optical flow estimation. In: Proceedings of the AAAI conference on artificial intelligence. vol.~31 (2017). \doi{10.1609/aaai.v31i1.10723}

\bibitem{sun2023scv3}
Sun, L., Bian, J.W., Zhan, H., Yin, W., Reid, I., Shen, C.: Sc-depthv3: Robust self-supervised monocular depth estimation for dynamic scenes. IEEE Transactions on Pattern Analysis and Machine Intelligence  \textbf{46}(1),  497--508 (2024). \doi{10.1109/TPAMI.2023.3322549}

\bibitem{Tukra2022Contrastive}
Tukra, S., Giannarou, S.: Stereo depth estimation via self-supervised contrastive representation learning. In: Wang, L., Dou, Q., Fletcher, P.T., Speidel, S., Li, S. (eds.) Medical Image Computing and Computer Assisted Intervention -- MICCAI 2022. pp. 604--614. Springer Nature Switzerland, Cham (2022)

\bibitem{Wang2022OMNET}
Wang, W., Ye, S., Wang, X., Zhao, Y.: Omnet: Real-time stereo matching with unsupervised occlusion mask. In: 2022 IEEE International Conference on Image Processing (ICIP). pp. 1241--1245 (2022). \doi{10.1109/ICIP46576.2022.9897748}

\bibitem{stereo2023Wei}
Wei, R., Li, B., Mo, H., Lu, B., Long, Y., Yang, B., Dou, Q., Liu, Y., Sun, D.: Stereo dense scene reconstruction and accurate localization for learning-based navigation of laparoscope in minimally invasive surgery. IEEE Transactions on Biomedical Engineering  \textbf{70}(2),  488--500 (2023). \doi{10.1109/TBME.2022.3195027}

\bibitem{Robust2022Xia}
Xia, W., Chen, E.C.S., Pautler, S., Peters, T.M.: A robust edge-preserving stereo matching method for laparoscopic images. IEEE Transactions on Medical Imaging  \textbf{41}(7),  1651--1664 (2022). \doi{10.1109/TMI.2022.3147414}

\bibitem{xu2023unimatch}
Xu, H., Zhang, J., Cai, J., Rezatofighi, H., Yu, F., Tao, D., Geiger, A.: Unifying flow, stereo and depth estimation. IEEE Transactions on Pattern Analysis and Machine Intelligence  \textbf{45}(11),  13941--13958 (2023). \doi{10.1109/TPAMI.2023.3298645}

\bibitem{Yan2019Segment}
Yan, T., Gan, Y., Xia, Z., Zhao, Q.: Segment-based disparity refinement with occlusion handling for stereo matching. IEEE Transactions on Image Processing  \textbf{28}(8),  3885--3897 (2019). \doi{10.1109/TIP.2019.2903318}

\bibitem{reconstruct2022Yang}
Yang, B., Xu, S., Chen, H., Zheng, W., Liu, C.: Reconstruct dynamic soft-tissue with stereo endoscope based on a single-layer network. IEEE Transactions on Image Processing  \textbf{31},  5828--5840 (2022). \doi{10.1109/TIP.2022.3202367}

\bibitem{yang2019hierarchical}
Yang, G., Manela, J., Happold, M., Ramanan, D.: Hierarchical deep stereo matching on high-resolution images. In: Proceedings of the IEEE/CVF Conference on Computer Vision and Pattern Recognition. pp. 5515--5524 (2019)

\bibitem{Yuan2024UnSAMFlow}
Yuan, S., Luo, L., Hui, Z., Pu, C., Xiang, X., Ranjan, R., Demandolx, D.: Unsamflow: Unsupervised optical flow guided by segment anything model. In: 2024 IEEE/CVF Conference on Computer Vision and Pattern Recognition (CVPR). pp. 19027--19037 (2024). \doi{10.1109/CVPR52733.2024.01800}

\bibitem{Zitnick2000LRC}
Zitnick, C., Kanade, T.: A cooperative algorithm for stereo matching and occlusion detection. IEEE Transactions on Pattern Analysis and Machine Intelligence  \textbf{22}(7),  675--684 (2000). \doi{10.1109/34.865184}

\end{thebibliography}

\end{document}